\pdfoutput=1

\documentclass[11pt]{article}

\usepackage{acl}

\usepackage{times}
\usepackage{latexsym}

\usepackage[T1]{fontenc}

\usepackage[utf8]{inputenc}
\usepackage{enumitem}
\usepackage{microtype}
\usepackage{amsmath}
\usepackage{graphicx}
\usepackage{multirow}
\usepackage{soul}

\usepackage{times}
\usepackage{latexsym}
\usepackage{graphicx}
\usepackage{booktabs}
\usepackage{subcaption}
\usepackage[ruled,vlined]{algorithm2e}
\usepackage{multirow}
\usepackage{tabularx}
\usepackage{amsmath}
\usepackage{bbold}
\usepackage{mathtools}

\usepackage{tabularx,ragged2e}
\usepackage{microtype}

\title{It's Better to Say ``I Can't Answer'' than Answering Incorrectly: Towards Selective Prediction in NLP}

\title{Towards Improving Selective Prediction Ability of NLP Systems}

\author{Neeraj Varshney,~~ 
  Swaroop Mishra,~~ 
  Chitta Baral
  \\
  Arizona State University \\
  \texttt{\{nvarshn2, srmishr1, cbaral\}}@asu.edu
  }

\begin{document}
\maketitle
\begin{abstract}

It's better to say ``I can't answer'' than to answer incorrectly.
This selective prediction ability is crucial for NLP systems to be reliably deployed in real-world applications.
Prior work has shown that existing selective prediction techniques fail to perform well, especially in the out-of-domain setting. 
In this work, we propose a method that improves probability estimates of models by calibrating them using prediction confidence and difficulty score of instances.
Using these two signals, we first annotate held-out instances and then train a calibrator to predict the likelihood of correctness of the model's prediction.
We instantiate our method with Natural Language Inference (NLI) and Duplicate Detection (DD) tasks and evaluate it in both In-Domain (IID) and Out-of-Domain (OOD) settings.
In (IID, OOD) settings, we show that the representations learned by our calibrator result in an improvement of $(15.81\%, 5.64\%)$ and $(6.19\%, 13.9\%)$ over \textit{MaxProb} --a selective prediction baseline-- on NLI and DD tasks respectively.

\end{abstract}

\section{Introduction}
\label{introduction}

In real-world applications, AI systems often encounter novel inputs that differ from their training data distribution.
Prior work has shown that even state-of-the-art models tend to make incorrect predictions on such inputs \cite{elsahar-galle-2019-annotate,miller2020effect,pmlr-v139-koh21a, hendrycks2021many}.
This raises reliability concerns and hinders their adoption in real-world safety-critical domains like biomedical and autonomous robots.
\textit{Selective prediction} addresses these concerns by enabling systems to abstain from making predictions when they are likely to be incorrect.
Avoiding incorrect predictions allows them to maintain high task accuracy and thus makes them more reliable.



\citet{hendrycks17baseline} proposed `\textit{MaxProb}' that uses the maximum softmax probability across all answer candidates as the confidence estimate to selectively make predictions.
While performing reasonably well in the \textit{in-domain} setting, \textit{MaxProb} and other existing selective prediction techniques fail to translate that performance in the \textit{out-of-domain} setting \cite{varshney2022investigating,kamath-etal-2020-selective}.

In this work, we propose a selective prediction method that improves probability estimates of models in both in-domain and out-of-domain settings by learning strong representations via calibration.
Specifically, we calibrate models' outputs using a held-out dataset and use the calibrator as confidence estimator for selective prediction.
To this end, we first argue that ``\textit{all instances are not equally difficult and the model is not equally confident in all its predictions''} and then through extensive experiments, we show that prediction confidence is positively correlated with correctness while difficulty score is negatively correlated (\ref{relationship_section}).
We leverage the above finding to calibrate models' outputs using these two signals.


For computing the difficulty scores, we use a \textit{model-based} technique (\ref{diff_computation_sec}) because human perception of difficulty may not always correlate well with machine interpretation.
To calibrate a model, we annotate instances of a held-out dataset conditioned on the model's predictive correctness (computed using difficulty score and prediction confidence) and then train a calibrator using these instances.
This annotation score represents the likelihood of correctness of the model's prediction.
Finally, the trained calibrator predicts this likelihood value for test instances and is used as the confidence estimator for selective prediction.





To evaluate the efficacy of our method, we conduct comprehensive experiments in In-Domain (IID) and Out-of-Domain (OOD) settings for Natural Language Inference (NLI) and Duplicate Detection (DD) tasks.
We also compare its performance with existing calibration techniques.
On the NLI task, our method achieves $15.81\%$ and $5.64\%$ improvement on AUC of \textit{risk-coverage} curve over \textit{MaxProb} in IID and OOD setting respectively.
Furthermore, on the DD task, it achieves  $6.19\%$ and $13.9\%$ improvement in IID and OOD setting respectively.
Finally, we hope that our work will facilitate development of more robust and reliable AI systems making their wide adoption in real-world applications possible.



\section{Selective Prediction}
Selective prediction enables a system to abstain on instances where it is likely to be incorrect i.e it consists of a \textit{selector} ($g$) that determines if the system should output the prediction.
Usually, $g$ comprises of a prediction confidence estimator $\tilde{g}$ and a threshold $th$ that controls the abstention level:
\begin{equation*}
    g(x) = \mathbb{1}[\tilde{g}(x)) > th]
\end{equation*}



A selective prediction system makes trade-offs between $coverage$ and $risk$.
For a dataset $D$, coverage at a threshold $th$ corresponds to the fraction of answered instances (where $\tilde{g} > th$) and risk is the error on those answered instances.

With the decrease in $th$, coverage will increase, but the risk will usually also increase.
The overall selective prediction performance across all thresholds is measured by the area under \textit{risk-coverage curve} \cite{el2010foundations}.
\textbf{Lower the AUC, the better the system} as it represents lower average risk across all thresholds.

\section{Method}

We propose to train a confidence estimator that can assign higher scores to correctly predicted instances than incorrectly predicted ones.
To this end, we leverage a held-out dataset and annotate it's instances conditioned on the model's predictive correctness.
Specifically, we infer the model on the held-out dataset and annotate instances with a score such that correctly predicted instances get assigned a higher score than incorrectly predicted instances.
This annotation score models the likelihood of the prediction being correct and is computed using the model's prediction confidence and difficulty level of the instance.
Finally, a calibrator (regression model) is trained using this annotated held-out dataset and used as the confidence estimator for selective prediction.

We detail each component of our method and the intuition behind it in the following subsections.

\subsection{Difficulty Score Computation}
\label{diff_computation_sec}
To compute difficulty score of an instance, we evaluate it after every training epoch and subtract the aggregated softmax probability assigned to the ground-truth answer from $1$ i.e. for an instance $i$, difficulty score $d_i$ is calculated as:
\[
     s_i = \frac{ \sum_{j=1}^{E} c_{ji}}{E} 
\]
\[
     d_i = 1 - s_i
\]

where the model is trained till $E$ epochs and $c_{ji}$ is prediction confidence of the correct answer given by the model after $j^{th}$ training epoch.
Note that $c_{ji}$ is probability assigned to the correct answer not the maximum probability across all answer candidates.
The intuition behind this procedure is that the \textit{instances that can be consistently answered correctly from the early stages of training are inherently easy and should receive lower difficulty score than the ones that require a large number of training steps.} 
A similar method has been explored in \citet{swayamdipta-etal-2020-dataset} for analyzing ``training dynamics'' but here we use it to quantify difficulty of the held-out instances.

\subsection{Annotation Score Computation}
We define annotation score for the held-out instances as a function of \emph{softmax probability} outputted by the model and the \emph{difficulty score}. 
We show that softmax score is positively correlated while difficulty score is negatively correlated with the predictive correctness i.e the system is more likely to be correct if the softmax score is high and difficulty score is low.
Furthermore, in order to justifiably separate the scores for correct and incorrect prediction scenarios in the range 0 to 1, we push the scores above 0.5 in case of correct and below 0.5 in case of incorrect scenarios. 
Concretely, we use the following functions to compute this:
\begin{equation*}
    {AS}_1 =
        \begin{cases}
          0.5+ \frac{maxProb}{2}, & \text{if correct} \\
          0.5- \frac{maxProb}{2}, & \text{otherwise}
        \end{cases}
\end{equation*}
\begin{equation*}
    {AS}_2 =
        \begin{cases}
          0.5+ \frac{s_i}{2}, & \text{if correct} \\
          0.5- \frac{s_i}{2}, & \text{otherwise}
        \end{cases}
\end{equation*}
\begin{equation*}
    {AS}_3 =
        \begin{cases}
          0.5+ \frac{max(s_i, maxProb)}{2}, & \text{if correct} \\
          0.5- \frac{min(s_i, maxProb)}{2}, & \text{otherwise}
        \end{cases}
\end{equation*}
${AS}_1$ uses only softmax, ${AS}_2$ uses only difficulty score and ${AS}_3$ uses a combination of both.
These annotation strategies assign a relatively higher score when the model's prediction is correct and a lower score when it is incorrect. 
This gold score ranges from 0 to 1 as both $s_i$ and $maxProb$ lie in the same range and better captures the likelihood of correctness unlike the categorical labels (1 for correct and 0 for incorrect) used in typical calibration approaches. 
\textbf{Note that this annotation computation is only required for training the calibrator and not at test time.} Therefore, difficulty score of the test instances need not be computed.

Both difficulty score and annotation score computation procedures are generic and are widely applicable since NLP systems usually make probabilistic predictions
for all kinds of tasks ranging from Classification to Question Answering. 

\subsection{Calibration}
\label{calib_sec}
Equipped with annotation scores, we extract syntactic features, namely, lengths, Semantic Textual Similarity (STS) value, number of common words between given sentences, and presence of negation words / numbers from the held-out instances to train the calibrator model. 
These features along with maxProb and prediction outputted by the model serve as inputs for the calibrator. 
Finally, we use a simple random forest implementation of Scikit-learn \cite{pedregosa2011scikit} to train our calibrator that learns strong representations for the inputs.
We note that these syntactic features are general and applicable for all language understanding tasks and any regression model can be used as the calibrator.
We compare our method with other calibration techniques described in Section \ref{baselines_sec}.

\section{Experimental Setup}


\subsection{Calibration Baselines}
\label{baselines_sec}
\citet{kamath-etal-2020-selective} study a calibration-based selective prediction technique for Question Answering datasets where they annotate a held-out dataset such that correctly predicted instances are assigned class label `1' and incorrect ones are assigned label `0'. 
Then, a calibrator is trained using this annotated binary classification dataset using features such as input length and probabilities of top 5 predictions.
The softmax probability assigned to class `1' by this calibrator is used as the confidence estimator for selective prediction.
We refer to this approach as \textbf{Calib C}.
We also train a transformer-based model for calibration (\textbf{Calib T}) that leverages the entire input text for this classification task instead of the syntactic features \cite{garg-moschitti-2021-will}.

Our proposed calibration method differs from these approaches as we quantify the correctness on a continuous scale (instead of categorical labels `1' and `0') using prediction confidence and difficulty of the instances and use explicitly provided general syntactic features described in Section \ref{calib_sec} for training. 
Our annotation procedure provides more flexibility for the calibrator to look for fine-grained features distinguishing various annotation scores. 
We note that our simplest annotation strategy (${AS}_1$) that does not incorporate difficulty score is similar to Calib R method described in \citet{varshney2022investigating} but our calibration method uses more general syntactic features.

Note that for fair estimation of abilities of the proposed method, we compare it with other calibration-based techniques only.
Other techniques such as Monte-Carlo dropout \cite{gal2016dropout} and Error Regularization \cite{xin-etal-2021-art} are complementary and can further improve our performance.

\subsection{Datasets}
We conduct experiments with Natural Language Inference and Duplicate Detection datasets and compare the performance of various calibration techniques in in-domain and out-of-domain settings.
\begin{table*}[]
\centering
\small
{%
\begin{tabular}{@{}l|l|lll|llll@{}}
\toprule
\multicolumn{1}{c}{\textbf{Method}} &
  \multicolumn{1}{c}{\textbf{{SNLI}}} &
  \multicolumn{3}{c}{\textbf{{MNLI}}} &
  \multicolumn{4}{c}{\textbf{{Stress Test}}} \\
  
  \multicolumn{1}{c}{} &
  \multicolumn{1}{l}{} &
  \textbf{Matched} & \textbf{Mismatched} & \textbf{Avg} &
  \textbf{Competence} & \textbf{Distraction} & \textbf{Noise} & \textbf{Avg}\\
 
\midrule
    \textbf{MaxProb (AUC)} & 2.78 & 14.00 &	14.44 &	14.22 & 47.87 &	26.49 &	20.34 &	31.57\\
    \midrule
    
    
    \textbf{Calib T (\%)} & -181.2 & -129.55 & -127.86 & -128.69 & -48.65 & -81.3 & -91.17 & -68.93\\
    \textbf{Calib C (\%)} & +8.97 & +2.15 & -1.36 & +0.40 & -3.75 & \textbf{+8.27} & -0.80 & +0.55 \\
    
    \textbf{Proposed (\%)} & \textbf{+15.81} & \textbf{+2.35} & \textbf{+2.04} & \textbf{+2.19} & \textbf{+8.01} & +6.60 & \textbf{+0.22} & \textbf{+5.64}\\


\bottomrule
\end{tabular}
}
\caption{Comparing percentage improvement of various calibration approaches on AUC of risk-coverage  curve (over MaxProb) in in-domain (SNLI) and out-of-domain settings (MNLI, Stress Test) for NLI task.}
\label{tab:NLI}

\end{table*}

\begin{table}[t]
    \centering
    \small
    \begin{tabular}{@{}l|cc@{}}
        \toprule
        \textbf{Method} & \textbf{MRPC} & \textbf{QQP} \\
        \midrule
        \textbf{MaxProb (AUC)} & 6.13 & 40.46 \\
        \midrule
        \textbf{Calib T (\%)}  & -148.87 & +2.21 \\
        \textbf{Calib C (\%)}  & -0.82 & +2.0 \\
        
        \textbf{Proposed (\%)}  & \textbf{+6.19} & \textbf{+13.9} \\

    \bottomrule
    \end{tabular}
    \caption{Comparing $\%$ improvement of various calibration approaches on AUC of risk-coverage curve in IID (MRPC) and OOD (QQP) settings for DD task.}
    \label{tab:DD}
\end{table}

\paragraph{NLI Datasets:}
SNLI \cite{bowman-etal-2015-large}, MNLI \cite{williams-etal-2018-broad} (Matched and Mismatched), and Stress Test \cite{naik-etal-2018-stress} (Competence, Distraction, and Noise).

\paragraph{Duplicate Detection Datasets:}
QQP \cite{iyer2017first} and MRPC \cite{dolan2005automatically}.

For NLI task, we train 3-way classification model (NLI has three labels) on SNLI and evaluate the selective prediction performance on SNLI (IID) and MNLI, Stress Test (OOD) datasets.
For the DD task, we train model on MRPC and evaluate on MRPC (IID) and QQP (OOD) datasets.
We use BERT-BASE model \cite{devlin-etal-2019-bert} with a linear layer on top of [CLS] token representation for training the model for these tasks.
We train these models with the default learning rate of $5e-5$ for $3$ epochs.\footnote{See Appendix for details}
We use the same experimental setup as \cite{varshney2022investigating} for calibration methods.


\section{Results and Analysis}


\subsection{MaxProb Struggles in OOD Setting}
First rows in Table \ref{tab:NLI} and \ref{tab:DD} show the AUC values achieved by MaxProb in NLI and DD tasks respectively. Note that in selective prediction, low AUC values of risk-coverage curves are preferred.
We find that MaxProb performs well in the IID setting as it achieves low AUC values ($2.78$ on SNLI and $6.13$ on MRPC).
However, it fails to translate that in the OOD setting (AUC of $14.22$ on MNLI, $31.57$ on Stress Test, and $40.46$ on QQP). This implies that the model makes a significant number of incorrect predictions with relatively high MaxProb and thus needs to be calibrated.

For calibration methods, we compare the performance improvement achieved over MaxProb w.r.t the minimum possible AUC.



\subsection{Proposed Method Outperforms All}
Our method shows a clear benefit over existing calibration techniques as it leads to a considerable improvement in all the cases. 
The proposed method achieves $15.81\%$ and $6.19\%$ improvement in the IID setting on SNLI and MRPC respectively. 
Furthermore, it achieves $2.19\%$ on MNLI, $5.64\%$ on Stress Test, and $13.9\%$ on QQP in the OOD setting.
\textit{Calib T considerably degrades performance in both IID and OOD settings.}
However, \textit{Calib C results in a minor improvement in the IID setting} ($8.97\%$ for SNLI) \textit{but does not consistently improve in the OOD setting} (especially on MNLI Mismatched and Competence Stress Test).
We attribute this to the limited signal that is given to the calibrator by annotating the held-out dataset with categorical labels `1' and `0'. Thus, it learns weak representations. 

\paragraph{Comparing Annotation Functions:}
We find that \textit{the improvement using our method comes from using ${AS}_3$ as the annotation score} which outperforms ${AS}_1$ and ${AS}_2$. 
This is expected as it leverages useful signals provided by both maxProb and difficulty score for annotation computation.

\begin{figure}
    \centering
    \includegraphics[width=4.5cm]{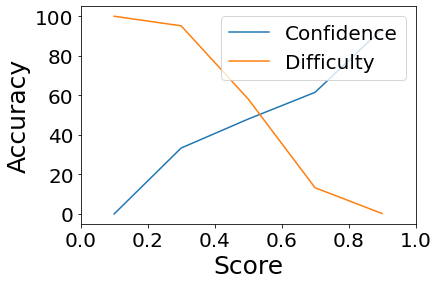}
    \caption{Trend of Model Accuracy with Confidence and Difficulty score for the NLI task.}
    \label{fig:relationship}
\end{figure}

\paragraph{Relationship With Predictive Correctness:}
To further analyze our method, we plot the relationship of predictive correctness with prediction confidence and difficulty score in Figure \ref{fig:relationship}.
It shows that prediction confidence is positively correlated while the difficulty score is negatively correlated with correctness. 
This further justifies our annotation score computation procedure.
\label{relationship_section}


\section{Conclusion and Future Work}

We proposed a selective prediction method that calibrates the model outputs using prediction confidence and difficulty level of the instances.
Through comprehensive experiments, we demonstrated that it achieves considerable improvement over MaxProb on NLI and Duplicate Detection tasks in both IID and OOD settings.
We hope that our work will facilitate development of more robust and reliable AI systems making their wide adoption in real-world applications possible.


\section*{Acknowledgements}
We thank the anonymous reviewers for their insightful feedback. 
This research was supported by DARPA SAIL-ON and DARPA CHESS programs.

\bibliography{anthology,custom}
\bibliographystyle{acl_natbib}

\appendix




\section*{Appendix}

\section{Related Work}

Instance-level difficulty analysis has recently received considerable attention.
\citet{varshney2022ildae} explore five different applications of difficulty analysis of evaluation data such as conducting efficient yet accurate evaluations with fewer instances and estimating OOD performance reliably.
\citet{rodriguez-etal-2021-evaluation} incorporate item response theory based difficulty quantification and analyze ranking reliability of leaderboards.
\citet{mishra2021robust} study robustness of model rankings by weighting instances based on their difficulty score.
\citet{swayamdipta-etal-2020-dataset} analyze the behavior of model on individual instances during training (\textit{training dynamics}) and categorize training instances into three different difficulty regions.


\section{Experimental Details}
We use batch size of $32$ on Nvidia V100 16GB GPUs for our experiments. 
We train these models with the default learning rate of $5e-5$ for $3$ epochs.
In Calib T approach, we use BERT-BASE model as the calibrator and train it using the annotated held-out dataset.
For training this calibrator, we use the default learning rate of $5e-5$. 
In the proposed approach, we use a simple random forest implementation of Scikit-learn \cite{pedregosa2011scikit} to train the calibrator.
Note that more advanced regression models could be used to further improve the performance of our approach. However, we leave that for future work as the focus of this paper is to show efficacy of our proposed approach on the selective prediction task.

\section{Features of Training Calibrator}
We extract syntactic features, namely, lengths, Semantic Textual Similarity (STS) value, number of common words between given sentences, and presence of negation words / numbers from the held-out instances to train the calibrator model. 
These features along with maxProb and prediction outputted by the model serve as inputs for the calibrator. 

For the NLI task, we compute these features for premise and hypothesis sentences i.e. STS value, number of common words, etc. between premise and hypothesis sentences.

Similarly, for the DD task, we compute these features for the given two sentences.



\end{document}